\documentclass[10pt,twocolumn,letterpaper]{article}

\usepackage[pagenumbers]{cvpr} 

\usepackage{graphicx}
\usepackage{amsmath}
\usepackage{amssymb}
\usepackage{booktabs}
\usepackage{dblfloatfix}

\usepackage{tikz}
\usepackage{comment}
\usepackage{dsfont}
\DeclareMathOperator*{\argmin}{argmin}
\usepackage{color}
\usepackage{soul} 
\usepackage{multirow}
\usepackage{subcaption}
\usepackage{pifont}
\usepackage{enumitem}
\newcommand{\cmark}{\ding{51}}%
\newcommand{\xmark}{\ding{55}}%

\newcommand{\encoder}[0]{query-based detection and tracking}

\newcommand{\ENCODER}[0]{Query-based Detection and Tracking}
\newcommand{\predictor}[0]{query-based prediction}

\newcommand{\PREDICTOR}[0]{Query-based Prediction}
\newcommand{\model}[0]{ViP3D}
\newcommand{\task}[0]{visual trajectory prediction task}

\newcommand{\inputs}[0]{raw videos}

\newcommand{\traditional}[0]{traditional perception and prediction pipeline}

\newcommand{\metric}[0]{EPA}

\newcommand{\baselineA}[0]{Agent trajectories}

\newcommand{\baselineD}[0]{Agent trajectories $+$ Agent queries}
\newcommand{\baselineAit}[0]{\textit{\baselineA}}

\newcommand{\baselineDit}[0]{\textit{\baselineD}}

\usepackage[pagebackref,breaklinks,colorlinks]{hyperref}

\usepackage[capitalize]{cleveref}
\crefname{section}{Sec.}{Secs.}
\Crefname{section}{Section}{Sections}
\Crefname{table}{Table}{Tables}
\crefname{table}{Tab.}{Tabs.}

\def\cvprPaperID{4686} 
\def\confName{CVPR}
\def\confYear{2023}

\begin{document}

\title{\model: End-to-end Visual Trajectory Prediction via 3D Agent Queries}

\author{
Junru Gu$\phantom{}^{1}\phantom{}^{*}$\hspace{15pt}
Chenxu Hu$\phantom{}^{1}\phantom{}^{*}$\hspace{15pt}
Tianyuan Zhang$\phantom{}^{2,3}$\hspace{15pt}
Xuanyao Chen$\phantom{}^{2,4}$
\and
Yilun Wang$\phantom{}^{5}$\hspace{15pt}
Yue Wang$\phantom{}^{6}$\hspace{15pt}
Hang Zhao$\phantom{}^{1,2}\phantom{}^{\text{†}}$ \vspace{10pt} \\
$\phantom{}^1$IIIS, Tsinghua University\hspace{5pt}
$\phantom{}^2$Shanghai Qi Zhi Institute \vspace{2pt}\\
$\phantom{}^3$CMU\hspace{5pt}
$\phantom{}^4$Fudan University\hspace{5pt}
$\phantom{}^5$Li Auto\hspace{5pt}
$\phantom{}^6$MIT\hspace{5pt} 
}

\twocolumn[{
\renewcommand\twocolumn[1][]{#1}
\maketitle
\vspace{-1.2cm}
\begin{center}
    \centering
    \includegraphics[width=1.0\linewidth]{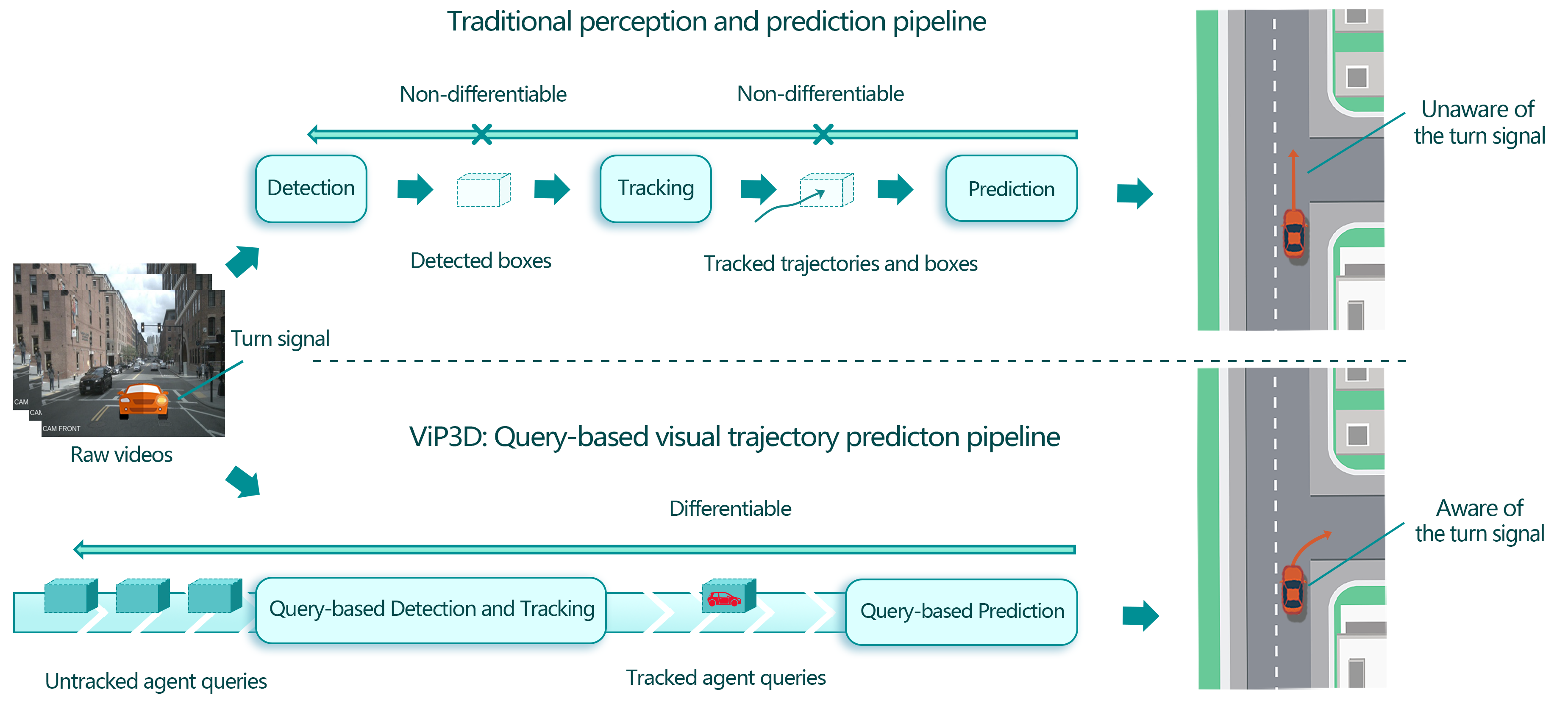} 
    \captionof{figure}{
    Comparison of a traditional multi-stage perception-prediction pipeline in autonomous driving and our proposed \model. The traditional pipeline involves multiple non-differentiable modules, \ie, detection, tracking, and prediction. \model\ uses 3D agent queries as the main thread of the pipeline, enabling end-to-end future trajectory prediction from raw video frame inputs. The novel design improves trajectory prediction performance by effectively leveraging fine-grained visual information such as the turning signals of vehicles.
         \label{fig:teaser}}
\end{center}
}]

\begin{abstract} 
{\begin{NoHyper}\let\thefootnote\relax\footnotetext{$^{*}$Equal contribution.}\end{NoHyper}}
{\begin{NoHyper}\let\thefootnote\relax\footnotetext{$^{\text{†}}$Corresponding to: hangzhao@mail.tsinghua.edu.cn}\end{NoHyper}}
Perception and prediction are two separate modules in the existing autonomous driving systems. They interact with each other via hand-picked features such as agent bounding boxes and trajectories.
Due to this separation, prediction, as a downstream module, only receives limited information from the perception module.
To make matters worse, errors from the perception modules can propagate and accumulate, adversely affecting the prediction results.
In this work, we propose \model, a query-based visual trajectory prediction pipeline that exploits rich information from raw videos to directly predict future trajectories of agents in a scene. \model\ employs sparse agent queries to detect, track, and predict throughout the pipeline, making it the first fully differentiable vision-based trajectory prediction approach.
Instead of using historical feature maps and trajectories, useful information from previous timestamps is encoded in agent queries, which makes \model\ a concise streaming prediction method.
Furthermore, extensive experimental results on the nuScenes dataset show the strong vision-based prediction performance of \model\ over traditional pipelines and previous end-to-end models.\footnote{Code and demos are available on the project page: \url{https://tsinghua-mars-lab.github.io/ViP3D}}
\end{abstract}

\vspace{-0.5cm}

\section{Introduction}
An autonomous driving system should be able to perceive agents in the current environment and predict their future behaviors so that the vehicle can navigate the world safely.
Perception and prediction are two separate modules in the existing autonomous driving software pipeline, where the interface between them is often defined as hand-picked geometric and semantic features, such as historical agent trajectories, agent types, agent sizes, \etc. Such an interface leads to the loss of useful perceptual information that can be used in trajectory prediction.
For example, tail lights and brake lights indicate a vehicle's intention, and pedestrians' head pose and body pose tell about their attention. This information, if not explicitly modeled, is ignored in the existing pipelines.
In addition, with the separation of perception and prediction, errors are accumulated and cannot be mitigated in later stages. Specifically, historical trajectories used by trajectory predictors come from an upstream perception module, which inevitably contains errors, leading to a drop in the prediction performance. Designing a trajectory predictor that is robust to upstream output errors is a non-trivial task~\cite{zhang2022trajectory}.

Recent works such as IntentNet~\cite{intentnet}, FaF~\cite{luo2018fast}, PnPNet~\cite{liang2020pnpnet} propose end-to-end models for LiDAR-based trajectory prediction. 
They suffer from a couple of limitations: (1) They are not able to leverage the abundant fine-grained visual information from cameras; (2) these models use convolutional feature maps as their intermediate representations within and across frames, thus suffering from non-differentiable operations such as non-maximum suppression in object decoding and object association in multi-object tracking.

To address all these challenges, we propose a novel pipeline that leverages a query-centric model design to predict future trajectories, dubbed \textbf{\model} (\textbf{Vi}sual trajectory \textbf{P}rediction via \textbf{3D} agent queries). \model\ consumes multi-view videos from surrounding cameras and high-definition maps, and makes agent-level future trajectory prediction in an end-to-end and concise streaming manner, as shown in Figure~\ref{fig:teaser}.
Specifically, \model\ leverages 3D agent queries as the interface throughout the pipeline, where each query can map to (at most) an agent in the environment.
At each time step, the queries aggregate visual features from multi-view images, learn agent temporal dynamics, model the relationship between agents, and finally produce possible future trajectories for each agent.
Across time, the 3D agent queries are maintained in a memory bank, which can be initialized, updated and discarded to track agents in the environment. 
Additionally, unlike previous prediction methods that utilize historical agent trajectories and feature maps from multiple historical frames, \model\ only uses 3D agent queries from one previous timestamp and sensor features from the current timestamp, making it a concise streaming approach.

In summary, the contribution of this paper is three-fold:
\begin{enumerate}[topsep=0pt,itemsep=-1ex,partopsep=1ex,parsep=1ex]
    \item \model\ is the first \textbf{fully differentiable vision-based} approach to predict future trajectories of agents for autonomous driving. Instead of using hand-picked features like historical trajectories and agent sizes, \model\ leverages the rich and fine-grained visual features from raw images which are useful for the trajectory prediction task.

    \item With \textbf{3D agent queries as interface}, \model\ explicitly models agent-level detection, tracking and prediction, making it interpretable and debuggable.

    \item \model\ is a concise model with \textbf{high performance}. It outperforms a wide variety of baselines and recent end-to-end methods on the visual trajectory prediction task.

\end{enumerate}

\section{Related Work}

\paragraph{3D Detection.}
There are a great number of works on 3D object detection and tracking from point clouds~\cite{pointnet,voxelnet,pointpillar}.
In this paper, we focus on 3D detection and tracking from cameras. Monodis~\cite{monodis} and FCOS3D~\cite{fcos3d} learn a single-stage object detector with instance depth and 3D pose predictions on monocular images. Pseudo-LiDAR~\cite{pseudoLiDAR} first predicts depth for each image pixel, then lifts them into the 3D space, and finally employs a point cloud based pipeline to perform 3D detection. DETR3D~\cite{detr3d} designs a sparse 3D query-based detection model that maps queries onto 2D multi-view images to extract features. BEVFormer~\cite{li2022bevformer} and PolarFormer~\cite{jiang2022polarformer} further propose a dense query-based detection model. Lift-Splat-Shoot~\cite{lift_splat_shoot} projects image features into BEV space by predicting depth distribution over pixels, BEVDet~\cite{huang2021bevdet} performs 3D object detection on top of it. Furthermore, PETR~\cite{liu2022petr} develops an implicit approach to transform 2D image features into BEV space for 3D detection.

\paragraph{3D Tracking.}
The majority of 3D tracking approaches follow the tracking-by-detection pipeline~\cite{am3dmot,simpletrack}. These methods first detect 3D objects, then associate existing tracklets with the new detections.
CenterTrack~\cite{centertrack,centerpoint} uses two consecutive frames to predict the speed of each detection box, then performs association using only $\ell_2$ distances of the boxes. Samuel~\etal~\cite{mono3dmot} uses PMBM filter to estimate states of tracklets and match them with new observations. DEFT~\cite{deft} uses a learned appearance matching network for association, together with an LSTM estimated motion to eliminate implausible trajectories. QD3DT~\cite{qd3dt} uses cues from depth-ordering and learns better appearance features via contrastive learning. MUTR3D~\cite{mutr3d} introduces track queries to model objects that appear in multiple cameras across multiple frames.

\paragraph{Trajectory Prediction.}
Several seminal trajectory prediction works have studied historical trajectory and map geometry encoding using graph neural networks~\cite{lanegcn,vectornet} and Transformers~\cite{SceneTransformer,nayakanti2022wayformer,varadarajan2022multipath++}.
To make multiple plausible future predictions~\cite{rasterize2019,phan2020covernet,fang2020tpnet,deo2018multi,phan2020covernet,multipath}, variety loss is a regression-based method that only optimizes the closest predicted trajectory during training. A Divide-And-Conquer~\cite{DivideAndConquer} approach is also a good initialization technique to produce diverse outputs. Modeling uncertainty using latent variables~\cite{cvae,hong2019rules,yeh2019diverse,sun2019stochastic,tang2019multiple,rhinehart2018r2p2,yuan2019diverse,Casas2020ImplicitLV,choi2022hierarchical} is another popular approach, which predicts different future trajectories by randomly sampling from the latent variables.
Goal-based methods recently achieve outstanding performance by first predicting the intentions of agents, such as the endpoint of trajectories~\cite{tnt,densetnt,gilles2021home,gilles2021thomas,2021goal}, lanes to follow~\cite{prime,kim2021lapred,lanegcn}, and then predicting trajectories conditioning on these goals.

\paragraph{End-to-End Perception and Prediction.}
In the last couple of years, there has been growing interest in jointly optimizing detection, tracking, and prediction.
FaF~\cite{luo2018fast} employs a single convolutional neural network to detect objects from LiDAR point clouds, and forecast their corresponding future trajectories.
IntentNet~\cite{intentnet} adds high-level intention output to this framework. More recently, Phillips~\etal~\cite{Phillips_2021_CVPR} further learns localization together with perception and prediction. FIERY~\cite{hu2021fiery} predicts future BEV occupancy heatmaps from visual data directly.
Mostly related to our work is PnPNet~\cite{liang2020pnpnet}, which explicitly models tracking in the loop.
Our method is related to these methods in the sense that we also perform end-to-end prediction based on sensor inputs. However, they all rely on BEV feature maps or heatmaps as their intermediate representation, which leads to unavoidable non-differentiable operation while going from dense feature maps to instance-level features, such as non-maximum suppression (NMS) in detection, and association in tracking. Our method, on the other hand, employs sparse agent queries as representation throughout the model, greatly improving the differentiability and interpretability.
\begin{figure*}[!t]
    \vspace{-0.7cm}
    \centering
    \includegraphics[width=0.99\linewidth]{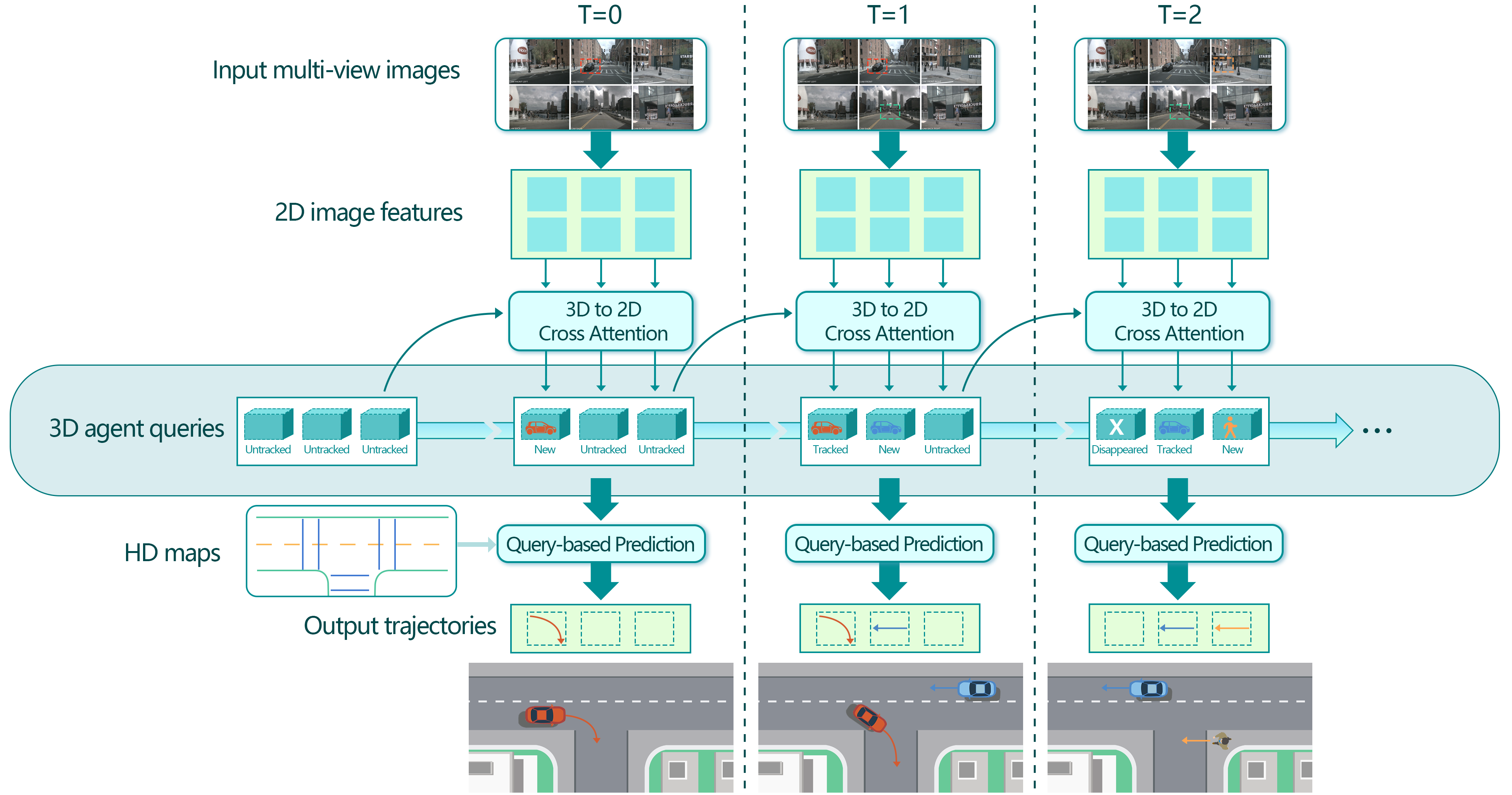}
    \caption{\model\ model pipeline. 3D agent queries serve as the main thread and intermediate representations over time. At each time step, the agent queries aggregate visual features from multi-view images to obtain tracked agent queries. The tracked queries further interact with HD maps and are decoded into predicted trajectories. The agent queries are managed in a dynamic memory bank, and the model works in a concise streaming manner.
    }
    \label{fig:arch}
    \vspace{-0.4cm}
\end{figure*}

\section{Method}

Overall, \model\ leverages a query-centric model design to address the trajectory prediction problem from \inputs\ in an end-to-end manner.
As shown in Figure~\ref{fig:arch}, 3D agent queries serve as the main thread across time. At each time step, a \encoder\ module extracts multi-view image features from surrounding cameras to update agent queries, forming a set of tracked agent queries. The tracked agent queries potentially contain much useful visual information, including the motion dynamics and visual characteristics of the agents.
After that, a \predictor\ module takes the tracked agent queries as input and associates them with HD map features, and finally outputs agent-wise future trajectories.
Over time, analogous to traditional trackers, the 3D agent queries are initialized, updated and discarded within a query memory bank, making \model\ work in a concise streaming fashion.
The design details of each module are explained in the following subsections.

\subsection{\ENCODER}
\label{sec:vis_enc}
For each input frame, a \encoder\ first extracts visual features from surrounding cameras, as shown in the upper part of Figure~\ref{fig:arch}.
Specifically,  we follow DETR3D~\cite{detr3d} to extract 2D features from multi-view images and use cross attention to update agent queries.
For temporal feature aggregation, inspired by MOTR~\cite{zeng2021motr}, we design a query-based tracking scheme with two key steps: query feature update and query supervision. Agent queries are updated across time to model the motion dynamics of agents.

\subsubsection{Query Feature Update}
Each agent query corresponds to at most one agent that appeared in the scene. We use $\mathbf{Q}$ to denote a set of agent queries, which are initialized as learnable embeddings with 3D reference points~\cite{detr3d}. 
At each time step, we first extract 2D image features of surrounding cameras via ResNet50~\cite{resnet} and FPN~\cite{fpn}. Then we project the 3D reference points of agent queries onto the 2D coordinates of multi-view images using camera intrinsic and extrinsic transformation matrices. Finally, we extract the corresponding image features $\mathbf{L}$ to update the agent queries via cross attention. Let $\mathbf{Q}_t'=\mathbf{Q}_t\mathbf{W}^Q, \mathbf{K} = \mathbf{L}\mathbf{W}^K, \mathbf{V} = \mathbf{L}\mathbf{W}^V$ be query / key / value vectors, respectively, where $\mathbf{W}^Q, \mathbf{W}^K, \mathbf{W}^V \in \mathbb{R}^{d_h\times d_k} $ are the matrices for linear projection, $t\in \{1, \dots, T\}$ is the current time step, $d_k$ is the dimension of query / key / value vectors. Then the cross attention is: $\tilde{\mathbf{Q}_t} = \operatorname{softmax} 
    \left(\frac{\mathbf{Q}_t'\mathbf{K}^\top}{\sqrt{d_k}}\right)\mathbf{V}$. Finally, we update the agent queries: $\mathbf{Q}'_{t} = \operatorname{FFN}\left( \mathbf{Q}_{t}+\tilde{\mathbf{Q}_t}\right)$, where $\rm{FFN}$ is a two-layer MLP with layer normalization.

\subsubsection{Query Supervision}
Since each agent query corresponds to at most one certain agent, supervision is required at each time step to make sure each query extracts features of the same agent across different historical frames. There are two types of queries. One is the matched queries that have been associated with ground truth agents before this time step. The other is the empty queries that have not been associated with any ground truth agent. Suppose we have done association at time step $t-1$, and now we perform association at time step $t$. For the matched queries, we assign the same ground truth agents to them as before:
$\mathbf{Q}_{\mathrm{matched}}\cong \mathcal{A}_{t-1}$,
where $\mathcal{A}_{t-1}$ denotes the ground truth agents at time step $t-1$.
If an agent disappears at time step $t$, we assign an empty label to supervise the corresponding agent query and reinitialize it as an empty unmatched query for later use. For the unmatched queries, we perform a bipartite matching between the unmatched queries and the new appeared agents $\mathcal{A}_{t,\mathrm{new}}$ at time step $t$:
$\mathbf{Q}_{\mathrm{empty}}\cong \mathcal{A}_{t,\mathrm{new}}$.

To perform the bipartite matching, we utilize a query decoder that outputs the center coordinates of each query at time step $t$. The pair-wise matching cost~\cite{detr} between ground truth $y_i$ and a prediction $\hat{y}_{\sigma(i)}$ for the bipartite matching is: $\mathcal{L}_{\rm{match}}\left(y_i, \hat{y}_{\sigma(i)} \right) = -\mathds{1}_{\left\{c_{i} \neq \varnothing\right\}} \hat{p}_{\sigma(i)}\left(c_{i}\right)+\mathds{1}_{\left\{c_{i} \neq \varnothing\right\}} \mathcal{L}_{\rm {box}}\left(b_{i}, \hat{b}_{\sigma(i)}\right)$,
where $c_i$ is the target class label, $\mathcal{L}_{\rm box}$ is the $\ell_1$ loss for bounding box parameters, $b_i$ is the target box, $\hat{b}_{\sigma(i)}$ and $\hat{p}_{\sigma(i)}\left(c_{i}\right)$ are the predicted box and predicted probability of class $c_i$, respectively.

After the bipartite matching, we get the optimal assignment $\hat{\sigma}$. We compute the query classification loss $\mathcal{L}_{\rm cls}$ and query coordinate regression loss $\mathcal{L}_{\rm coord}$ as follows:
\vspace{-0.2cm}
\begin{align}
\mathcal{L}_{\rm cls} &= \sum_{i=1}^{N}-\log \hat{p}_{\hat{\sigma}(i)}\left(c_{i}\right), \\
\mathcal{L}_{\rm coord} &= \sum_{i=1}^{N} \mathds{1}_{\left\{c_{i} \neq \varnothing\right\}} \mathcal{L}_{\rm box}\left(b_{i}, \hat{b}_{\hat{\sigma}}(i)\right),
\end{align}
\vspace{-0.1cm}
where $\mathcal{L}_{\rm box}$ is the $\ell_1$ loss for bounding box parameters.

\subsubsection{Query Memory Bank}
To model long-term relationships for agent queries of different time steps, we maintain historical states for each agent query in a query memory bank. Following MOTR~\cite{zeng2021motr}, the memory query bank is a first-in-first-out queue with a fixed size $S_{\rm bank}$. After each time step, the attention mechanism is only applied between each query and its historical states in the memory bank for efficiency. For the $i^{th}$ agent query $q^{i}_{t}$ at the time step $t$, the corresponding historical states in the memory bank are denoted as $\mathbf{Q}^{i}_{\rm bank}=\{{q}^{i}_{t-S_{\rm bank}}, \dots ,{q}^{i}_{t-2},{q}^{i}_{t-1}\}$. Then the temporal cross attention is $\tilde{q^{i}_t} = \operatorname{softmax} 
    \left(\frac{q^{i}_{t,{\rm query}}{\mathbf{Q}^{i}_{\rm bank,key}}^\top}{\sqrt{d}}\right)\mathbf{Q}^{i}_{\rm bank,value}$, where $q^{i}_{t,{\rm query}}$, $\mathbf{Q}^{i}_{\rm bank,key}$, $\mathbf{Q}^{i}_{\rm bank,value}$ are query / key / value vectors after linear projection, respectively, and d is the dimension of the agent queries.
The $i^{th}$ agent query is updated by: ${q^{i}_t}' = \operatorname{FFN}\left( q^{i}_t+\tilde{q^{i}_t}\right)$, where $\rm{FFN}$ is a two-layer MLP with layer normalization. Finally, the historical states of the $i^{th}$ agent query in the memory bank become: ${\mathbf{Q}^{i}_{\rm bank}}'=\{{q}^{i}_{t-S_{\rm bank}+1}, \dots ,{q}^{i}_{t-1},{{q}^{i}_{t}}'\}$.

\subsection{\PREDICTOR}
Typical trajectory prediction models can be divided into three components: an agent encoder that extracts agent trajectory features, a map encoder that extracts map features, and a trajectory decoder that outputs predicted trajectories.
In our pipeline, the \encoder\ gives tracked agent queries, which is equivalent to the output of the agent encoder. Therefore, by taking agent queries as input, the \predictor\ module is composed of only a map encoder and a trajectory decoder.

\subsubsection{Map Encoding}
HD semantic maps are crucial for trajectory prediction since they include detailed road information, such as lane types, road boundaries, and traffic signs. HD maps are typically represented by vectorized spatial coordinates of map elements and the topological relations between them. To encode this information, we adopt a popular vectorized encoding method VectorNet~\cite{vectornet}.
The map encoder produces a set of map features $\mathbf{M}$, which further interacts with agent queries via cross attention: $\mathbf{Q}' = \operatorname{Attention}(\mathbf{Q},\mathbf{M})$.

\subsubsection{Trajectory Decoding}
The trajectory decoding takes the agent queries as input and outputs $K$ possible future trajectories for each agent.
\model\ is compatible with a variety of trajectory decoding methods, such as regression-based methods~\cite{social-gan,cui2019multimodal,rupprecht2017ambiguity,lanegcn}, goal-based methods~\cite{tnt} and heatmap-based methods~\cite{densetnt,gilles2021home,gilles2021gohome}. 
We introduce the key ideas of these methods here and leave the details in the Appendix.
(1) The regression-based method, namely variety loss (or min-of-K), predicts future trajectories based on regression. During inference, this decoder directly outputs a set of predicted trajectories. During training, we first calculate the distance between each predicted trajectory and the ground truth trajectory. Then we select a predicted trajectory with the closest distance and only calculate regression loss between it and the ground truth trajectory.
(2) The goal-based method first defines sparse goal anchors heuristically and then classifies these anchors to estimate and select the goals. Finally, a trajectory is completed for each selected goal.
(3) The heatmap-based method first generates a heatmap indicating the probability distribution of the goal. Then a greedy algorithm or a neural network is used to select goals from the heatmap. Finally, same as the goal-based method, the trajectories are completed. We use $\mathcal{L}_{\rm traj}$ to denote the loss of trajectory decoding and leave the detailed definition in the Appendix.

\subsection{Loss}
\model\ is trained end-to-end with query classification loss and query coordinate regression loss of the \encoder, and trajectory decoding loss of the \predictor: $\mathcal{L} = \mathcal{L}_{\rm cls} + \mathcal{L}_{\rm coord} + \mathcal{L}_{\rm traj}$.

\section{Experiments}

%

\subsection{End-to-end Prediction Accuracy}
\begin{figure}[t]
    \centering
    \includegraphics[width=\columnwidth]{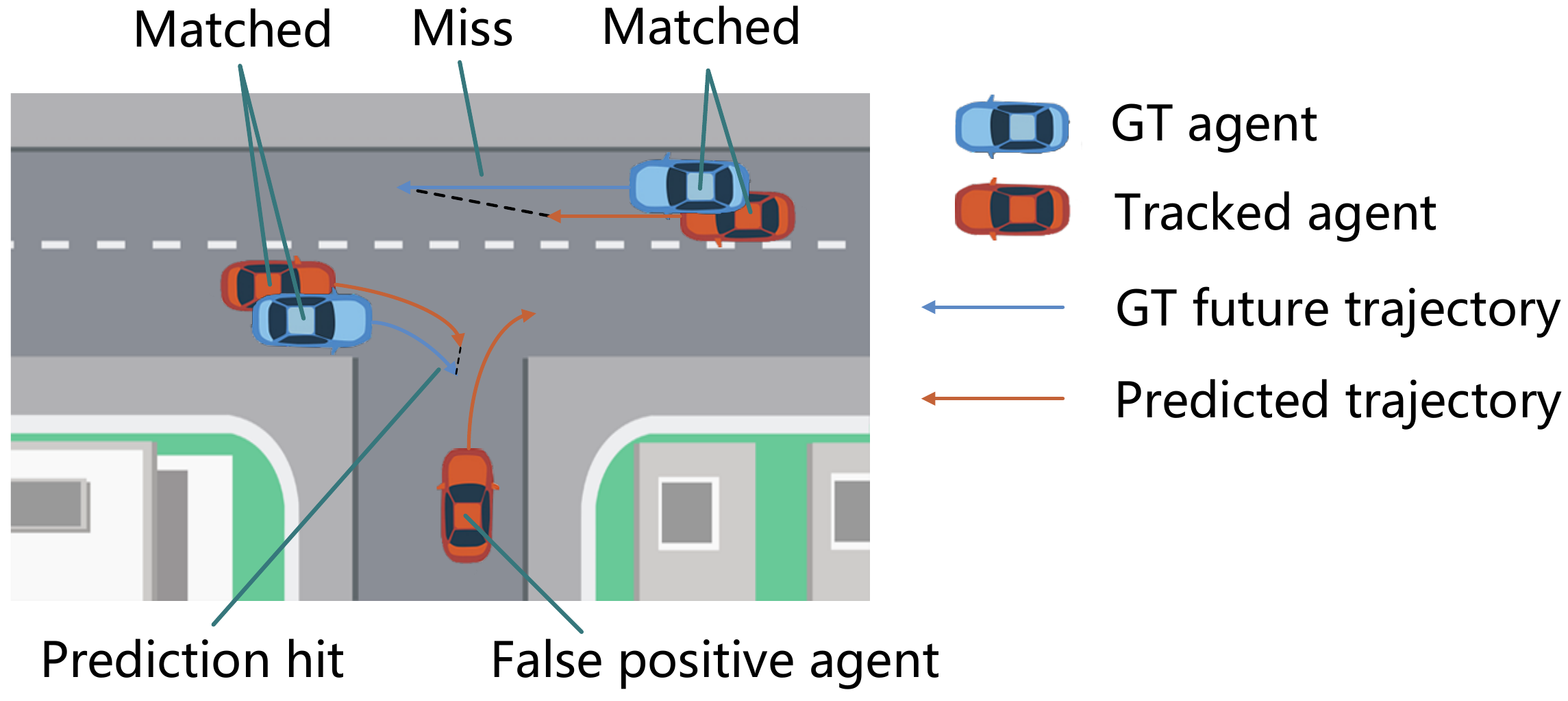}
    \caption{An example of End-to-end Prediction Accuracy (EPA) calculation. Blue and red agents are ground truth and detected agents, respectively. After matching the ground truth and the detection results, the red agent in the lower part is considered a false positive agent. A predicted trajectory is considered a hit when its final displacement error is below a certain threshold.}
    \label{fig:metric}
\end{figure}

To evaluate the performance of multi-future trajectory prediction, we adopt the common metrics including minimum average displacement error (minADE), minimum final displacement error (minFDE), and miss rate (MR).
However, the inputs of end-to-end prediction are raw pixels, models may detect more false positive agents which should not exist (an example shown in Figure~\ref{fig:metric}). In these metrics, we find the closest predicted trajectory for each ground truth trajectory to calculate displacement error, which does not account for false positives. Therefore, we propose a more comprehensive evaluation metric for end-to-end visual trajectory prediction, named End-to-end Prediction Accuracy (EPA).

Let us denote predicted and ground truth agents as unordered sets $ \hat{\mathcal{S}}$ and $\mathcal{S}$, respectively, where each agent is represented by $K$ future trajectories of different modalities.
First, for each agent type $c$, we calculate the prediction precision between $\hat{\mathcal{S}}_c$ and $\mathcal{S}_c$, where the subscript $c$ indicates the agents of type $c$.
We define the cost between a predicted agent $\hat{\mathbf{s}}$ and a ground truth agent $\mathbf{s}$ as:
\vspace{-0.2cm}
\begin{equation}
    C_{\mathrm{EPA}}(\mathbf{s},\hat{\mathbf{s}}) = \begin{cases}
    ||s_{0}-\hat{s}_{0}||,&\text{if } ||s_{0}-\hat{s}_{0}||\leqslant  \tau _{\mathrm{EPA}}\\
    \infty, &\text{if } ||s_{0}-\hat{s}_{0}||>\tau_{\mathrm{EPA}}
    \end{cases},
\end{equation} 
where $\hat{s}_0$ and $s_0$ indicate the coordinates of the ground truth agent and the predicted agent at the current time step, and we set the threshold of successful matching to $\tau _{\mathrm{EPA}}=2.0\mathrm{m}$.
We utilize bipartite matching according to $C_{\mathrm{EPA}}$ to find the correspondence between predicted agents and ground truth agents. Then the number of false-positive predicted agents is $N_{\mathrm{FP}}=|\hat{\mathcal{S}}|-|\hat{\mathcal{S}}_{\mathrm{match}}|$,
where $\hat{\mathcal{S}}_{\mathrm{match}} \subset \hat{\mathcal{S}}$ is the set of predicted agents which have been matched with ground truth agents. For each matched agent, we calculate $\mathrm{minFDE}$ (minimum final displacement error) between its predicted multiple future trajectories and the ground truth trajectory $\mathrm{minFDE}(\hat{\mathbf{s}},\mathbf{s})=\min\limits_{k\in 1\dots K} ||\hat{s}^{(k)}_{T_{\mathrm{future}}}-s_{T_{\mathrm{future}}}||$,
where $\hat{\mathbf{s}}^{(k)}$ is the $k^{th}$ trajectory of the matched agent $\hat{\mathbf{s}}$ , and $T_{\mathrm{future}}$ is the final time step of the future trajectory. Now the set of agents which have matched and hit a ground truth agent is $\hat{\mathcal{S}}_{\mathrm{match, hit}}=\{\hat{\mathbf{s}}:\hat{\mathbf{s}}\in \hat{\mathcal{S}}_{\mathrm{match}},\mathrm{minFDE}(\hat{\mathbf{s}}, \mathbf{s})\leqslant \tau _{\mathrm{EPA}} \}$.
The \metric\ between $\hat{\mathcal{S}}_c$ and $\mathcal{S}_c$ is defined as:
\begin{equation}
    \mathrm{\metric}(\hat{\mathcal{S}}_c, \mathcal{S}_c) =
    \dfrac{|\hat{\mathcal{S}}_{\mathrm{match, hit}}|-\alpha N_{\mathrm{FP}}}{N_{\mathrm{GT}}},
\end{equation}
where $N_{\mathrm{GT}}$ is the number of ground truth agents, and we set the penalty coefficient $\alpha=0.5$ for all experiments. For different scenes, each number in the equation is defined as the sum over all scenes. Finally, the \metric\ between $\hat{\mathcal{S}}$ and $\mathcal{S}$ is averaged over all agent types.

\begin{table*}[t!]
\small
\centering
\begin{tabular}{llccccc}
\hline
                               &                                                  & \multicolumn{2}{c}{Traditional}      & \multicolumn{2}{c}{PnPNet-vision~\cite{liang2020pnpnet}} & ViP3D (Ours)     \\ \hline
\multirow{5}{*}{Architechture} & \multicolumn{1}{l|}{detector}                    & \multicolumn{2}{c}{DETR3D}           & \multicolumn{2}{c}{DETR3D}                                 & DETR3D           \\
                               & \multicolumn{1}{l|}{detector-tracker interface}  & \multicolumn{2}{c}{boxes}            & \multicolumn{2}{c}{boxes}                                  & queries          \\
                               & \multicolumn{1}{l|}{tracker}                     & Kalman Filter      & CenterPoint     & Kalman Filter                 & CenterPoint                & query-based      \\
                               & \multicolumn{1}{l|}{tracker-predictor interface} & \multicolumn{2}{c}{trajectories}     & \multicolumn{2}{c}{cropped features}                       & queries          \\
                               & \multicolumn{1}{l|}{predictor}                   & \multicolumn{2}{c}{regression-based} & \multicolumn{2}{c}{regression-based}                         & regression-based   \\ \hline
\multirow{4}{*}{Metrics}       & \multicolumn{1}{l|}{minADE$\downarrow$}          & 2.07               & 2.06            & 2.04                          & 2.04                       & \textbf{2.03}  \\
                               & \multicolumn{1}{l|}{minFDE$\downarrow$}          & 3.10               & 3.02            & 3.08                          & 3.03                       & \textbf{2.90}  \\
                               & \multicolumn{1}{l|}{MR$\downarrow$}              & 0.289              & 0.277           & 0.277                         & 0.271                      & \textbf{0.239} \\
                               & \multicolumn{1}{l|}{EPA$\uparrow$}               & 0.191              & 0.209           & 0.198                         & 0.213                      & \textbf{0.236} \\ \hline
\end{tabular}
\caption{Comparing \model\ with traditional multi-stage pipeline. Classical metrics include minADE, minFDE and Miss Rate (MR), and End-to-end Prediction Accuracy (EPA) which is our proposed metric for the end-to-end setting. For each agent, 6 future trajectories with a time horizon of 6 seconds are evaluated.
\label{table:main_result}
}
\end{table*}

\subsection{Experimental Settings}

\paragraph{Dataset.}
We train and evaluate \model\ on the nuScenes dataset, a large-scale driving dataset including the urban scenarios in Boston and Singapore. It contains 1000 scenes, and each scene has a duration of around 20 seconds. The full dataset has more than one million images from 6 cameras and 1.4M bounding boxes for different types of objects. Bounding boxes of objects are annotated at 2Hz over the entire dataset. 

\paragraph{Trajectory Prediction Settings.}
Popular trajectory prediction benchmarks, such as Argoverse Motion Prediction Benchmark~\cite{argoverse}, require the prediction of one target agent in each scene. 
In our \task, we simultaneously predict all agents in each scene, which is the same as real-time usage. A commonly used trick is to predict trajectories in allocentric view, \ie, taking the last position of the target agent as the origin and its direction as $y$-axis. It makes prediction models focus on future modality prediction instead of coordinate transformation, thereby improving the prediction performance. In our experiments, we use this trick for all baselines and our \model. Metrics averaged over vehicles and pedestrians are used to compare their performance on \task.

\subsection{Baseline Settings}
\label{sec:baseline_setting}
\paragraph{Traditional Perception and Prediction Pipeline.}
The traditional pipeline is composed of a vision-based detector, a tracker, and a predictor. For a fair comparison, the vision-based detector is the same as \model. For the tracker, we test the performance of the classical IoU association with Kalman Filter, and an advanced tracking method named CenterPoint~\cite{centerpoint}. Compared with \model, the outputs of the tracker are agent trajectories and agent attributes instead of agent queries. These agent attributes are manually-defined in common tracking tasks, and we use as many attributes as possible, including agent types, agent sizes, agent velocities, \textit{etc}.

\paragraph{PnPNet-vision.}
PnPNet~\cite{liang2020pnpnet} only takes LiDAR data as input, and it cannot be directly used for our \task. Following the original PnPNet, we propose \textit{PnPNet-vision} by replacing the LiDAR encoder of the original PnPNet with DETR3D, which is the same as the detector of \model. Instead of using the query-based tracker and predictor, PnPNet associates boxes across frames according to affinity matrix and uses Kalman Filter as the motion model, which is a non-differentiable operation. For prediction, PnPNet crops features from the BEV feature map according to tracked trajectories, and takes the cropped features as the inputs of the prediction. We use Lift-Splat-Shot to obtain the BEV feature map for PnPNet-vision.

\subsection{Evaluation and Analysis}

\subsubsection{Main Results}
We compare our \model\ with \traditional\ and PnPNet-vision on the nuScenes dataset, as shown in Table~\ref{table:main_result}. The \traditional\ uses historical trajectories as the interface between tracking and prediction, so it cannot utilize visual information for prediction. Our proposed PnPNet-vision follows the key idea of the original PnPNet to obtain agent features by cropping from BEV feature maps, and takes the cropped features as the inputs of the predictor. More implementation details are described in Section~\ref{sec:baseline_setting}.
All baselines and our \model\ use DETR3D as the detector and regression-based trajectory decoding method as the predictor for a fair comparison. 
We can see that \model\ outperforms these baselines on all the metrics, indicating the effectiveness and superiority of directly learning from visual information with a fully differentiable approach.

\begin{table*}[]
\centering
\small
\begin{tabular}{@{}l|c|c|cccc@{}}
\toprule
             & {Prediction inputs} & {Differentiable} & {minADE $\downarrow$}& {minFDE $\downarrow$} & {MR $\downarrow$} & {\metric $\uparrow$} \\ \midrule
             & Agent trajectories                                & \xmark                           &         2.30                             &      3.33                                &             0.282                         & 0.186                                 \\
             & Agent trajectories + Agent queries                & \xmark                           &           2.20                           &            3.19                          &                   0.274                    & 0.211                                 \\ \midrule
\model\      & Agent queries                                     & \cmark                           &         \bf  2.03                        &           \bf2.90                     &                \bf 0.239                      & \textbf{0.236}                        \\
\bottomrule
\end{tabular}
\caption{
Ablation study on the inputs of the trajectory prediction module of \model. Trajectory decoding defaults to a regression-based method.
\label{table:ablation_study}
}
\end{table*}

\subsubsection{Ablation Study}
\paragraph{Trajectory Prediction Inputs.}
To better understand the necessity of visual features and end-to-end training, we compare \model\ with different baselines. These baselines have the same architecture as \model\ except for the prediction inputs.
We use the default regression-based method for trajectory decoding.
Results are shown in Table~\ref{table:ablation_study}. It can be seen that \baselineDit\ outperforms \baselineAit, demonstrating that the agent queries provide more fine-grained and detailed visual information to improve prediction performance. \model\ surpasses \baselineAit\ and \baselineDit, demonstrating that fully differentiable end-to-end learning is helpful in avoiding the error accumulation problem in the multi-stage pipeline.

\paragraph{Trajectory Decoding Methods.}
We compare our \model\ with \traditional\ under other trajectory decoding methods, goal-based TNT~\cite{tnt} and heatmap-based HOME~\cite{gilles2021home}, which recently achieve state-of-the-art performance. 
As shown in Table~\ref{table:traj_dec_result}, \model\ surpasses the \traditional\ on these metrics under the two trajectory decoding methods, demonstrating that \model\ is compatible with various state-of-the-art trajectory decoders and achieves superior performance.

\begin{table}[h]
\small
\centering
\begin{tabular}{l|l|cccc}
\toprule
\multirow{1}{*}{Decoder}                 & \multirow{1}{*}{Pipeline}             &  mADE  &mFDE  & MR  & EPA  \\
\midrule
\multirow{2}{*}{Goal~\cite{tnt}}               & Traditional                           &     2.50             &      3.93          &   0.266         &    0.195       \\
                                                     & \model\                         &  \bf2.24             &   \bf3.33          & \bf0.238        & \bf0.219       \\
\midrule
\multirow{2}{*}{Heatmap~\cite{gilles2021home}} & Traditional                           &     2.53             &      3.81          &   0.264         &    0.197       \\
                                                     & \model\                         &  \bf2.33             &   \bf3.42          & \bf0.218        & \bf0.214       \\
\bottomrule
\end{tabular}
\vspace{-0.3cm}
\caption{Comparing trajectory prediction performance on the nuScenes validation set with another two trajectory decoding methods: goal-based and heatmap-based. mADE and mFDE denote minADE and minFDE, respectively.
\label{table:traj_dec_result}
}
\end{table}

\paragraph{View of Trajectory Prediction.}
We test the performance of the pipelines in two different prediction coordinates. One is in the egocentric view, and the other is in the allocentric view~\cite{jia22a}. The egocentric view indicates predicting trajectories in the coordinate system of the ego vehicle, while the allocentric view indicates predicting trajectories in the coordinate system of the predicted agent itself. Predicting trajectories in the allocentric view is a commonly used normalization trick, and it has a better performance compared with the egocentric view. As shown in Table~\ref{table:prediction_view}, the same results are obtained in our experiments. So experiments of baselines and \model\ in other sections are performed in the allocentric view by default.

\vspace{-0.4cm} 
\paragraph{Analysis of Different Detectors}
We also conduct experiments on other vision-based detectors, such as PETRv2~\cite{liu2022petrv2}, which leverages the temporal information of previous frames to assist 3D object detection. When using PETRv2 as the detection backbone, \model\ achieves a better performance in short-term inference ($<3\mathrm{s}$) but fails in long-term inference ($>  10\mathrm{s}$). It indicates that the performance of long-term inference is sensitive to the detection backbone, and more efforts are needed to adapt \model\ to different detectors. A possible solution is to run \model\ on longer scene segments (currently 3 frames) during training if the GPU memory is large enough. We regard it as a limitation of \model.

\begin{table}[]
\centering
\small
\begin{tabular}{@{}l|l|cccc@{}}
\toprule
View                         & Pipeline    & minADE          & minFDE          & MR               & EPA              \\ \midrule
\multirow{2}{*}{Egocentric}  & Traditional & 2.51            & 3.57            & 0.353            & 0.132            \\
                             & ViP3D       & 2.10            & 3.01            & 0.261            & 0.199            \\ \midrule
\multirow{2}{*}{Allocentric} & Traditional & 2.06            & 3.02            & 0.277            & 0.209            \\
                             & ViP3D       & $\textbf{2.03}$ & $\textbf{2.90}$ & $\textbf{0.239}$ & $\textbf{0.236}$ \\ \bottomrule
\end{tabular}
\caption{
The comparison between different types of view of trajectory prediction.
\label{table:prediction_view}
}
\vspace{-0.5cm} 
\end{table}

\begin{figure*}[!b]
    \centering
    \vspace{-0.5cm} 
    \includegraphics[width=0.9\linewidth]{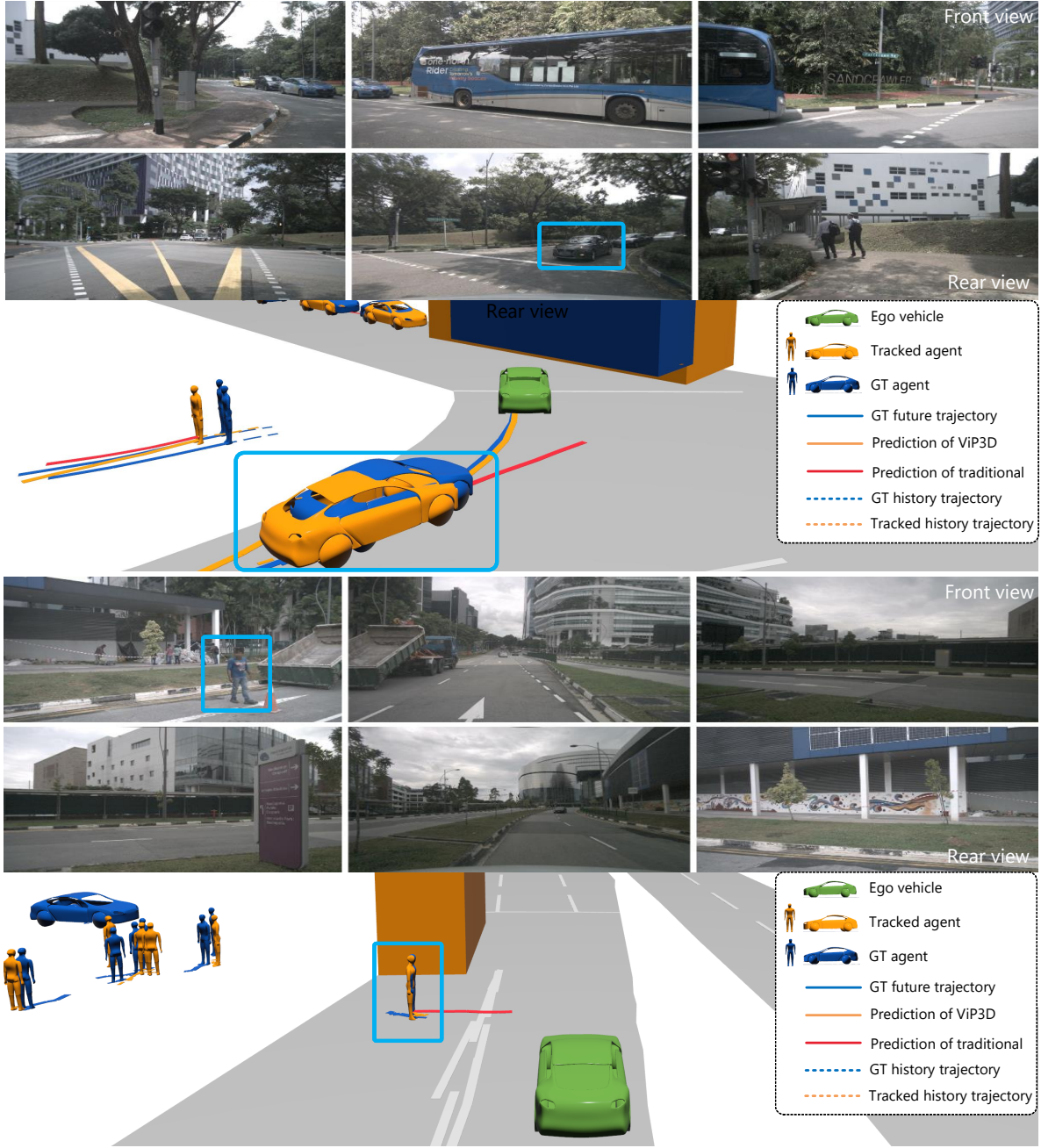}
    \vspace{-0.3cm}
    \caption{Qualitative results. Input camera images are shown on the top. The green vehicle is the ego agent. The blue and orange agents indicate ground-truth and tracked agents, respectively. The blue, orange and red curves indicate ground-truth trajectories, prediction of ViP3D and prediction of the traditional pipeline, respectively. For each agent, only the predicted trajectory with the highest probability is drawn.}
    \label{fig:vis}
    \vspace{-1.0cm}  
\end{figure*}

\vspace{-0.4cm}   
\subsubsection{Qualitative Results}
We provide examples of the predicted results by \model\ and traditional pipeline in Figure~\ref{fig:vis}.
In the upper example, we can see that the left turn signal of the vehicle in the blue box is flashing, indicating that the vehicle is about to turn left. ViP3D can use this visual information to predict the correct trajectory. In contrast, the traditional pipeline can only use historical trajectory information to predict that the vehicle is about to go straight incorrectly.
In the lower example, we can see that the pedestrian is facing the coming vehicle, indicating that he has probably noticed the approaching vehicle and will stop and wait for the vehicle to go first. ViP3D makes use of the pedestrian's head pose to correctly predict that the pedestrian will stop, while the traditional pipeline incorrectly predicts that pedestrians will cross the road.
These two examples show that ViP3D improves trajectory prediction performance due to utilizing visual information.

\vspace{-0.13cm} 
\section{Conclusion}
\vspace{-0.1cm}
We present \model, a fully differentiable approach to predict future trajectories of agents from multi-view videos. It exploits the rich visual information from the raw sensory input and avoids the error accumulation problem in the traditional pipeline.
Moreover, by leveraging 3D agent queries, \model\ models agent instances explicitly, making the pipeline interpretable and debuggable. 

\clearpage

{\small
\bibliographystyle{ieee_fullname}
\bibliography{example}

\begin{thebibliography}{10}\itemsep=-1pt

\bibitem{detr}
Nicolas Carion, Francisco Massa, Gabriel Synnaeve, Nicolas Usunier, Alexander
  Kirillov, and Sergey Zagoruyko.
\newblock {End-to-end object detection with transformers}.
\newblock In {\em ECCV}, 2020.

\bibitem{Casas2020ImplicitLV}
S. Casas, Cole Gulino, Simon Suo, Katie Luo, Renjie Liao, and R. Urtasun.
\newblock Implicit latent variable model for scene-consistent motion
  forecasting.
\newblock In {\em ECCV}, 2020.

\bibitem{intentnet}
Sergio Casas, Wenjie Luo, and Raquel Urtasun.
\newblock Intentnet: Learning to predict intention from raw sensor data.
\newblock In {\em Conference on Robot Learning}, pages 947--956. PMLR, 2018.

\bibitem{deft}
Mohamed Chaabane, Peter Zhang, J~Ross Beveridge, and Stephen O'Hara.
\newblock Deft: Detection embeddings for tracking.
\newblock {\em arXiv preprint arXiv:2102.02267}, 2021.

\bibitem{multipath}
Yuning Chai, Benjamin Sapp, Mayank Bansal, and Dragomir Anguelov.
\newblock Multipath: Multiple probabilistic anchor trajectory hypotheses for
  behavior prediction.
\newblock {\em arXiv preprint arXiv:1910.05449}, 2019.

\bibitem{argoverse}
Ming-Fang Chang, John Lambert, Patsorn Sangkloy, Jagjeet Singh, Slawomir Bak,
  Andrew Hartnett, De Wang, Peter Carr, Simon Lucey, Deva Ramanan, et~al.
\newblock Argoverse: 3d tracking and forecasting with rich maps.
\newblock In {\em Proceedings of the IEEE/CVF Conference on Computer Vision and
  Pattern Recognition}, pages 8748--8757, 2019.

\bibitem{choi2022hierarchical}
Dooseop Choi and KyoungWook Min.
\newblock Hierarchical latent structure for multi-modal vehicle trajectory
  forecasting.
\newblock In {\em Computer Vision--ECCV 2022: 17th European Conference, Tel
  Aviv, Israel, October 23--27, 2022, Proceedings, Part XXII}, pages 129--145.
  Springer, 2022.

\bibitem{rasterize2019}
Henggang Cui, Vladan Radosavljevic, Fang-Chieh Chou, Tsung-Han Lin, Thi Nguyen,
  Tzu-Kuo Huang, Jeff Schneider, and Nemanja Djuric.
\newblock Multimodal trajectory predictions for autonomous driving using deep
  convolutional networks.
\newblock In {\em 2019 International Conference on Robotics and Automation
  (ICRA)}, pages 2090--2096. IEEE, 2019.

\bibitem{cui2019multimodal}
Henggang Cui, Vladan Radosavljevic, Fang-Chieh Chou, Tsung-Han Lin, Thi Nguyen,
  Tzu-Kuo Huang, Jeff Schneider, and Nemanja Djuric.
\newblock Multimodal trajectory predictions for autonomous driving using deep
  convolutional networks.
\newblock In {\em 2019 International Conference on Robotics and Automation
  (ICRA)}, pages 2090--2096. IEEE, 2019.

\bibitem{adam}
Kingma Da.
\newblock A method for stochastic optimization.
\newblock {\em arXiv preprint arXiv:1412.6980}, 2014.

\bibitem{deo2018multi}
Nachiket Deo and Mohan~M Trivedi.
\newblock Multi-modal trajectory prediction of surrounding vehicles with
  maneuver based lstms.
\newblock In {\em 2018 IEEE Intelligent Vehicles Symposium (IV)}, pages
  1179--1184. IEEE, 2018.

\bibitem{fang2020tpnet}
Liangji Fang, Qinhong Jiang, Jianping Shi, and Bolei Zhou.
\newblock Tpnet: Trajectory proposal network for motion prediction.
\newblock In {\em Proceedings of the IEEE/CVF Conference on Computer Vision and
  Pattern Recognition}, pages 6797--6806, 2020.

\bibitem{vectornet}
Jiyang Gao, Chen Sun, Hang Zhao, Yi Shen, Dragomir Anguelov, Congcong Li, and
  Cordelia Schmid.
\newblock Vectornet: Encoding hd maps and agent dynamics from vectorized
  representation.
\newblock In {\em Proceedings of the IEEE/CVF Conference on Computer Vision and
  Pattern Recognition}, pages 11525--11533, 2020.

\bibitem{gilles2021gohome}
Thomas Gilles, Stefano Sabatini, Dzmitry Tsishkou, Bogdan Stanciulescu, and
  Fabien Moutarde.
\newblock Gohome: Graph-oriented heatmap output for future motion estimation.
\newblock {\em arXiv preprint arXiv:2109.01827}, 2021.

\bibitem{gilles2021home}
Thomas Gilles, Stefano Sabatini, Dzmitry Tsishkou, Bogdan Stanciulescu, and
  Fabien Moutarde.
\newblock Home: Heatmap output for future motion estimation.
\newblock {\em arXiv preprint arXiv:2105.10968}, 2021.

\bibitem{gilles2021thomas}
Thomas Gilles, Stefano Sabatini, Dzmitry Tsishkou, Bogdan Stanciulescu, and
  Fabien Moutarde.
\newblock Thomas: Trajectory heatmap output with learned multi-agent sampling.
\newblock In {\em International Conference on Learning Representations}, 2021.

\bibitem{densetnt}
Junru Gu, Chen Sun, and Hang Zhao.
\newblock Densetnt: End-to-end trajectory prediction from dense goal sets.
\newblock In {\em Proceedings of the IEEE/CVF International Conference on
  Computer Vision}, pages 15303--15312, 2021.

\bibitem{social-gan}
Agrim Gupta, Justin Johnson, Li Fei-Fei, Silvio Savarese, and Alexandre Alahi.
\newblock Social gan: Socially acceptable trajectories with generative
  adversarial networks.
\newblock In {\em Proceedings of the IEEE Conference on Computer Vision and
  Pattern Recognition}, pages 2255--2264, 2018.

\bibitem{resnet}
Kaiming He, Xiangyu Zhang, Shaoqing Ren, and Jian Sun.
\newblock {Deep Residual Learning for Image Recognition}.
\newblock In {\em CVPR}, pages 770--778, 2016.

\bibitem{hong2019rules}
Joey Hong, Benjamin Sapp, and James Philbin.
\newblock Rules of the road: Predicting driving behavior with a convolutional
  model of semantic interactions.
\newblock In {\em Proceedings of the IEEE/CVF Conference on Computer Vision and
  Pattern Recognition}, pages 8454--8462, 2019.

\bibitem{hu2021fiery}
Anthony Hu, Zak Murez, Nikhil Mohan, Sof{\'\i}a Dudas, Jeffrey Hawke, Vijay
  Badrinarayanan, Roberto Cipolla, and Alex Kendall.
\newblock Fiery: Future instance prediction in bird's-eye view from surround
  monocular cameras.
\newblock In {\em Proceedings of the IEEE/CVF International Conference on
  Computer Vision}, pages 15273--15282, 2021.

\bibitem{qd3dt}
Hou-Ning Hu, Yung-Hsu Yang, Tobias Fischer, Trevor Darrell, Fisher Yu, and Min
  Sun.
\newblock Monocular quasi-dense 3d object tracking.
\newblock {\em arXiv preprint arXiv:2103.07351}, 2021.

\bibitem{huang2021bevdet}
Junjie Huang, Guan Huang, Zheng Zhu, and Dalong Du.
\newblock Bevdet: High-performance multi-camera 3d object detection in
  bird-eye-view.
\newblock {\em arXiv preprint arXiv:2112.11790}, 2021.

\bibitem{jia22a}
Xiaosong Jia, Liting Sun, Hang Zhao, Masayoshi Tomizuka, and Wei Zhan.
\newblock Multi-agent trajectory prediction by combining egocentric and
  allocentric views.
\newblock In Aleksandra Faust, David Hsu, and Gerhard Neumann, editors, {\em
  Proceedings of the 5th Conference on Robot Learning}, volume 164 of {\em
  Proceedings of Machine Learning Research}, pages 1434--1443. PMLR, 08--11 Nov
  2022.

\bibitem{jiang2022polarformer}
Yanqin Jiang, Li Zhang, Zhenwei Miao, Xiatian Zhu, Jin Gao, Weiming Hu, and
  Yu-Gang Jiang.
\newblock Polarformer: Multi-camera 3d object detection with polar
  transformers.
\newblock {\em arXiv preprint arXiv:2206.15398}, 2022.

\bibitem{kim2021lapred}
ByeoungDo Kim, Seong~Hyeon Park, Seokhwan Lee, Elbek Khoshimjonov, Dongsuk Kum,
  Junsoo Kim, Jeong~Soo Kim, and Jun~Won Choi.
\newblock {LaPred}: Lane-aware prediction of multi-modal future trajectories of
  dynamic agents.
\newblock In {\em Proceedings of the IEEE/CVF Conference on Computer Vision and
  Pattern Recognition}, pages 14636--14645, 2021.

\bibitem{pointpillar}
Alex~H Lang, Sourabh Vora, Holger Caesar, Lubing Zhou, Jiong Yang, and Oscar
  Beijbom.
\newblock {PointPillars: Fast Encoders for Object Detection from Point Clouds}.
\newblock In {\em CVPR}, pages 12697--12705, 2019.

\bibitem{cvae}
Namhoon Lee, Wongun Choi, Paul Vernaza, Christopher~B Choy, Philip~HS Torr, and
  Manmohan Chandraker.
\newblock Desire: Distant future prediction in dynamic scenes with interacting
  agents.
\newblock In {\em Proceedings of the IEEE Conference on Computer Vision and
  Pattern Recognition}, pages 336--345, 2017.

\bibitem{li2022bevformer}
Zhiqi Li, Wenhai Wang, Hongyang Li, Enze Xie, Chonghao Sima, Tong Lu, Qiao Yu,
  and Jifeng Dai.
\newblock Bevformer: Learning bird's-eye-view representation from multi-camera
  images via spatiotemporal transformers.
\newblock {\em arXiv preprint arXiv:2203.17270}, 2022.

\bibitem{lanegcn}
Ming Liang, Bin Yang, Rui Hu, Yun Chen, Renjie Liao, Song Feng, and Raquel
  Urtasun.
\newblock Learning lane graph representations for motion forecasting.
\newblock In {\em European Conference on Computer Vision}, pages 541--556.
  Springer, 2020.

\bibitem{liang2020pnpnet}
Ming Liang, Bin Yang, Wenyuan Zeng, Yun Chen, Rui Hu, Sergio Casas, and Raquel
  Urtasun.
\newblock Pnpnet: End-to-end perception and prediction with tracking in the
  loop.
\newblock In {\em Proceedings of the IEEE/CVF Conference on Computer Vision and
  Pattern Recognition}, pages 11553--11562, 2020.

\bibitem{fpn}
Tsung-Yi Lin, Piotr Doll{\'a}r, Ross Girshick, Kaiming He, Bharath Hariharan,
  and Serge Belongie.
\newblock {Feature Pyramid Networks for Object Detection}.
\newblock In {\em CVPR}, pages 2117--2125, 2017.

\bibitem{liu2022petr}
Yingfei Liu, Tiancai Wang, Xiangyu Zhang, and Jian Sun.
\newblock Petr: Position embedding transformation for multi-view 3d object
  detection.
\newblock {\em arXiv preprint arXiv:2203.05625}, 2022.

\bibitem{liu2022petrv2}
Yingfei Liu, Junjie Yan, Fan Jia, Shuailin Li, Qi Gao, Tiancai Wang, Xiangyu
  Zhang, and Jian Sun.
\newblock Petrv2: A unified framework for 3d perception from multi-camera
  images.
\newblock {\em arXiv preprint arXiv:2206.01256}, 2022.

\bibitem{luo2018fast}
Wenjie Luo, Bin Yang, and Raquel Urtasun.
\newblock Fast and furious: Real time end-to-end 3d detection, tracking and
  motion forecasting with a single convolutional net.
\newblock In {\em Proceedings of the IEEE conference on Computer Vision and
  Pattern Recognition}, pages 3569--3577, 2018.

\bibitem{DivideAndConquer}
Sriram Narayanan, Ramin Moslemi, Francesco Pittaluga, Buyu Liu, and Manmohan
  Chandraker.
\newblock Divide-and-conquer for lane-aware diverse trajectory prediction.
\newblock In {\em Proceedings of the IEEE/CVF Conference on Computer Vision and
  Pattern Recognition}, pages 15799--15808, 2021.

\bibitem{nayakanti2022wayformer}
Nigamaa Nayakanti, Rami Al-Rfou, Aurick Zhou, Kratarth Goel, Khaled~S Refaat,
  and Benjamin Sapp.
\newblock Wayformer: Motion forecasting via simple \& efficient attention
  networks.
\newblock {\em arXiv preprint arXiv:2207.05844}, 2022.

\bibitem{SceneTransformer}
Jiquan Ngiam, Benjamin Caine, Vijay Vasudevan, Zhengdong Zhang, Hao-Tien~Lewis
  Chiang, Jeffrey Ling, Rebecca Roelofs, Alex Bewley, Chenxi Liu, Ashish
  Venugopal, et~al.
\newblock Scene transformer: A unified multi-task model for behavior prediction
  and planning.
\newblock {\em arXiv preprint arXiv:2106.08417}, 2021.

\bibitem{simpletrack}
Ziqi Pang, Zhichao Li, and Naiyan Wang.
\newblock Simpletrack: Understanding and rethinking 3d multi-object tracking.
\newblock {\em arXiv preprint arXiv:2111.09621}, 2021.

\bibitem{phan2020covernet}
Tung Phan-Minh, Elena~Corina Grigore, Freddy~A Boulton, Oscar Beijbom, and
  Eric~M Wolff.
\newblock Covernet: Multimodal behavior prediction using trajectory sets.
\newblock In {\em Proceedings of the IEEE/CVF Conference on Computer Vision and
  Pattern Recognition}, pages 14074--14083, 2020.

\bibitem{lift_splat_shoot}
Jonah Philion and Sanja Fidler.
\newblock Lift, splat, shoot: Encoding images from arbitrary camera rigs by
  implicitly unprojecting to 3d.
\newblock In {\em European Conference on Computer Vision}, pages 194--210.
  Springer, 2020.

\bibitem{Phillips_2021_CVPR}
John Phillips, Julieta Martinez, Ioan~Andrei Barsan, Sergio Casas, Abbas Sadat,
  and Raquel Urtasun.
\newblock Deep multi-task learning for joint localization, perception, and
  prediction.
\newblock In {\em Proceedings of the IEEE/CVF Conference on Computer Vision and
  Pattern Recognition (CVPR)}, pages 4679--4689, June 2021.

\bibitem{pointnet}
Charles~R Qi, Hao Su, Kaichun Mo, and Leonidas~J Guibas.
\newblock {PointNet: Deep Learning on Point Sets for 3D Classification and
  Segmentation}.
\newblock In {\em CVPR}, pages 652--660, 2017.

\bibitem{rhinehart2018r2p2}
Nicholas Rhinehart, Kris~M Kitani, and Paul Vernaza.
\newblock R2p2: A reparameterized pushforward policy for diverse, precise
  generative path forecasting.
\newblock In {\em Proceedings of the European Conference on Computer Vision
  (ECCV)}, pages 772--788, 2018.

\bibitem{rupprecht2017ambiguity}
Christian Rupprecht, Iro Laina, Robert DiPietro, Maximilian Baust, Federico
  Tombari, Nassir Navab, and Gregory~D Hager.
\newblock Learning in an uncertain world: Representing ambiguity through
  multiple hypotheses.
\newblock In {\em Proceedings of the IEEE international conference on computer
  vision}, pages 3591--3600, 2017.

\bibitem{mono3dmot}
Samuel Scheidegger, Joachim Benjaminsson, Emil Rosenberg, Amrit Krishnan, and
  Karl Granstr{\"o}m.
\newblock Mono-camera 3d multi-object tracking using deep learning detections
  and pmbm filtering.
\newblock In {\em 2018 IEEE Intelligent Vehicles Symposium (IV)}, pages
  433--440. IEEE, 2018.

\bibitem{monodis}
Andrea Simonelli, Samuel~Rota Bulo, Lorenzo Porzi, Manuel L{\'o}pez-Antequera,
  and Peter Kontschieder.
\newblock Disentangling monocular 3d object detection.
\newblock In {\em Proceedings of the IEEE/CVF International Conference on
  Computer Vision}, pages 1991--1999, 2019.

\bibitem{prime}
Haoran Song, Di Luan, Wenchao Ding, Michael~Yu Wang, and Qifeng Chen.
\newblock Learning to predict vehicle trajectories with model-based planning.
\newblock {\em arXiv preprint arXiv:2103.04027}, 2021.

\bibitem{sun2019stochastic}
Chen Sun, Per Karlsson, Jiajun Wu, Joshua~B Tenenbaum, and Kevin Murphy.
\newblock Stochastic prediction of multi-agent interactions from partial
  observations.
\newblock {\em arXiv preprint arXiv:1902.09641}, 2019.

\bibitem{tang2019multiple}
Yichuan~Charlie Tang and Ruslan Salakhutdinov.
\newblock Multiple futures prediction.
\newblock {\em arXiv preprint arXiv:1911.00997}, 2019.

\bibitem{2021goal}
Hung Tran, Vuong Le, and Truyen Tran.
\newblock Goal-driven long-term trajectory prediction.
\newblock In {\em Proceedings of the IEEE/CVF Winter Conference on Applications
  of Computer Vision}, pages 796--805, 2021.

\bibitem{varadarajan2022multipath++}
Balakrishnan Varadarajan, Ahmed Hefny, Avikalp Srivastava, Khaled~S Refaat,
  Nigamaa Nayakanti, Andre Cornman, Kan Chen, Bertrand Douillard, Chi~Pang Lam,
  Dragomir Anguelov, et~al.
\newblock Multipath++: Efficient information fusion and trajectory aggregation
  for behavior prediction.
\newblock In {\em 2022 International Conference on Robotics and Automation
  (ICRA)}, pages 7814--7821. IEEE, 2022.

\bibitem{fcos3d}
Tai Wang, Xinge Zhu, Jiangmiao Pang, and Dahua Lin.
\newblock Fcos3d: Fully convolutional one-stage monocular 3d object detection.
\newblock {\em arXiv preprint arXiv:2104.10956}, 2021.

\bibitem{pseudoLiDAR}
Yan Wang, Wei-Lun Chao, Divyansh Garg, Bharath Hariharan, Mark Campbell, and
  Kilian~Q Weinberger.
\newblock {Pseudo-LiDAR from Visual Depth Estimation: Bridging the Gap in 3D
  Object Detection for Autonomous Driving}.
\newblock In {\em CVPR}, pages 8445--8453, 2019.

\bibitem{detr3d}
Yue Wang, Vitor~Campagnolo Guizilini, Tianyuan Zhang, Yilun Wang, Hang Zhao,
  and Justin Solomon.
\newblock Detr3d: 3d object detection from multi-view images via 3d-to-2d
  queries.
\newblock In {\em 5th Annual Conference on Robot Learning}, 2021.

\bibitem{am3dmot}
Xinshuo Weng, Jianren Wang, David Held, and Kris Kitani.
\newblock {3D Multi-Object Tracking: A Baseline and New Evaluation Metrics}.
\newblock {\em IROS}, 2020.

\bibitem{yeh2019diverse}
Raymond~A Yeh, Alexander~G Schwing, Jonathan Huang, and Kevin Murphy.
\newblock Diverse generation for multi-agent sports games.
\newblock In {\em Proceedings of the IEEE/CVF Conference on Computer Vision and
  Pattern Recognition}, pages 4610--4619, 2019.

\bibitem{centerpoint}
Tianwei Yin, Xingyi Zhou, and Philipp Kr{\"a}henb{\"u}hl.
\newblock {Center-based 3D Object Detection and Tracking}.
\newblock {\em arXiv preprint arXiv:2006.11275}, 2020.

\bibitem{yuan2019diverse}
Ye Yuan and Kris~M Kitani.
\newblock Diverse trajectory forecasting with determinantal point processes.
\newblock In {\em International Conference on Learning Representations}, 2019.

\bibitem{zeng2021motr}
Fangao Zeng, Bin Dong, Tiancai Wang, Xiangyu Zhang, and Yichen Wei.
\newblock Motr: End-to-end multiple-object tracking with transformer.
\newblock {\em arXiv preprint arXiv:2105.03247}, 2021.

\bibitem{zhang2022trajectory}
Pu Zhang, Lei Bai, Jianru Xue, Jianwu Fang, Nanning Zheng, and Wanli Ouyang.
\newblock Trajectory forecasting from detection with uncertainty-aware motion
  encoding.
\newblock {\em arXiv preprint arXiv:2202.01478}, 2022.

\bibitem{mutr3d}
Tianyuan Zhang, Xuanyao Chen, Yue Wang, Yilun Wang, and Hang Zhao.
\newblock Mutr3d: A multi-camera tracking framework via 3d-to-2d queries.
\newblock In {\em Proceedings of the IEEE/CVF Conference on Computer Vision and
  Pattern Recognition}, pages 4537--4546, 2022.

\bibitem{tnt}
Hang Zhao, Jiyang Gao, Tian Lan, Chen Sun, Benjamin Sapp, Balakrishnan
  Varadarajan, Yue Shen, Yi Shen, Yuning Chai, Cordelia Schmid, et~al.
\newblock Tnt: Target-driven trajectory prediction.
\newblock {\em arXiv preprint arXiv:2008.08294}, 2020.

\bibitem{centertrack}
Xingyi Zhou, Vladlen Koltun, and Philipp Kr{\"a}henb{\"u}hl.
\newblock Tracking objects as points.
\newblock In {\em European Conference on Computer Vision}, pages 474--490.
  Springer, 2020.

\bibitem{voxelnet}
Yin Zhou and Oncel Tuzel.
\newblock {VoxelNet: End-to-End Learning for Point Cloud Based 3D Object
  Detection}.
\newblock In {\em CVPR}, pages 4490--4499, 2018.

\end{thebibliography}
}

\clearpage
\appendix

\section{Implementation Details}
\paragraph{Training and Inference Details.}
In our experiments, all models are trained on the nuScenes training set with a batch size of 8 for 24 epochs. The ADAM optimizer~\cite{adam} is adopted to train the whole pipeline. The learning rate has an initial value of $2e^{-4}$ and decays to 10\% at the 20th and the 23rd epochs. The hidden size of the \encoder\ module is set to 256, and that of the trajectory predictor is set to 128. A pretrained detection backbone is used for model initialization.
We evaluate all models on the nuScenes validation set. All models are tested online by feeding raw multi-view images of each time step to the model in chronological order. The metric computation is performed at every step except for steps that do not have enough future frames. Different from popular trajectory prediction benchmarks that only require predictions of selected agents, we simultaneously predict all agents at each step.

\paragraph{\ENCODER.}
The \encoder\ takes ResNet50~\cite{resnet} as the image backbone and DETR3D~\cite{detr3d} as the detection head. The detection head consists of 6 layers, and each layer contains a feature refinement layer and a multi-head attention layer with layer normalization. The hidden size for the detection head is set to 256. Finally, one branch predicts center coordinates and size of agents, and the other branch predicts agent type. Each branch consists of two fully connected layers, where the hidden size is also 256. 

\paragraph{Map Encoding.}
Same as typical trajectory prediction models, \model\ also encodes HD maps to facilitate trajectory prediction. VectorNet~\cite{vectornet} is the first trajectory prediction method to encode vectorized HD maps using a hierarchical graph neural network, and we follow it to convert each lane into a sequence of vectors. Each vector represents a segment of the lane, including the endpoints of the segment, the attributes of the lane, and the numerical order of the segment in the lane. 



\section{Trajectory Decoding}
\model\ can leverage a variety of trajectory decoding methods, such as regression-based methods~\cite{social-gan,cui2019multimodal,rupprecht2017ambiguity,lanegcn}, goal-based methods~\cite{tnt} and heatmap-based methods~\cite{densetnt,gilles2021home,gilles2021gohome}. We conduct experiments on these three trajectory decoding methods. In this section, we introduce the implementation details of these methods.

\paragraph{Regression-based.}
The regression-based trajectory decoder is a 2-layer MLP that takes the agent queries as input and directly outputs multiple future trajectories. During inference, the regression-based trajectory decoder directly outputs a set of predicted trajectories. During training, we first calculate the distance between each predicted trajectory $\hat{\mathbf{s}}$ and ground truth trajectory $\mathbf{s}$:
$d(\mathbf{s}, \hat{\mathbf{s}})=\sum\limits_{t=1}^{T_{\mathrm{future}}} ||s_{t}-\hat{s}_{t}||,$
where $||\cdot||$ is the $\ell_2$ distance between two points. Then, we select the predicted trajectory with the closest distance:
$\hat{k} = \mathop{\argmin}_{ k\in 1\dots K } { d(\mathbf{s}, {\mathbf{s}}^{(k)}) },$
where ${\mathbf{s}}^{(k)}$ is the $k^{th}$ predicted trajectory. Finally, we calculate regression loss between the closest predicted trajectory ${\mathbf{s}}^{(\hat{k})}$ and the ground truth trajectory $\mathbf{s}$ as 
\begin{equation}
    \mathcal{L}_{\rm trajectory} = \sum\limits_{t=1}^{T_{\mathrm{future}}}\mathcal{L}_{\rm reg}(s_{t},s^{(\hat{k})}_{t}),
\end{equation}
where $\mathcal{L}_{\rm reg}$ is the smooth $\ell_1$ loss between two points.

\paragraph{Goal-based.}
The goal-based trajectory decoder consists of a goal encoder, a probability decoder, an offset decoder, and a trajectory completion module. These modules are implemented using MLP. For each agent, we first randomly generate a set of candidate goals. The goal encoder is used to obtain the features of candidate goals by taking their coordinates as input. After that, a concatenation of the agent query and the features of goal coordinates is fed into the probability decoder and offset decoder. The probability decoder and the offset decoder output predicted goal probabilities and goal offsets, respectively. Let $\mathcal{L}_{\rm cls}$ be the binary cross-entropy loss for the probability decoder, and let $\mathcal{L}_{\rm reg}$ be the smooth $\ell_1$ loss for the offset decoder. To obtain $K$ trajectories, Non-maximum supervision (NMS) is employed to select $K$ goals (after adding the goal offsets), and the trajectory completion module takes the $K$ selected goals and outputs $K$ trajectories. Let $\mathcal{L}_{\rm completion}$ be the smooth $\ell_1$ loss for the trajectory completion module. Then the overall loss is 
\begin{equation}
    \mathcal{L}_{\rm trajectory} = \mathcal{L}_{\rm cls} + \mathcal{L}_{\rm reg} + \mathcal{L}_{\rm completion}.
\end{equation}

\paragraph{Heatmap-based.}
The heatmap-based trajectory decoder only consists of a goal encoder, a probability decoder, and a trajectory completion module. These modules are implemented using MLP. For each agent, to obtain a heatmap indicating the probability distribution of the final positions of the trajectories, we first densely sample goals with a sampling density of 1m. The goal encoder is used to obtain the features of the goals by taking their coordinates as input. After that, a concatenation of the agent query and the features of goal coordinates is fed into the probability decoder. The probability decoder outputs predicted goal probabilities, and we obtain the heatmap. Let $\mathcal{L}_{\rm cls}$ be the binary cross-entropy loss for the probability decoder. To obtain $K$ trajectories, we also use NMS to select $K$ goals for simplification, instead of using greedy algorithms as in origin heatmap-based methods~\cite{gilles2021home}. The trajectory completion module takes the $K$ selected goals and outputs $K$ trajectories. Let $\mathcal{L}_{\rm completion}$ be the smooth $\ell_1$ loss for the trajectory completion module. Then the overall loss is 
\begin{equation}
    \mathcal{L}_{\rm trajectory} = \mathcal{L}_{\rm cls} + \mathcal{L}_{\rm completion}.
\end{equation}

\section{Qualitative Results}
\begin{figure*}[!b]
    \centering
    \includegraphics[width=0.9\linewidth]{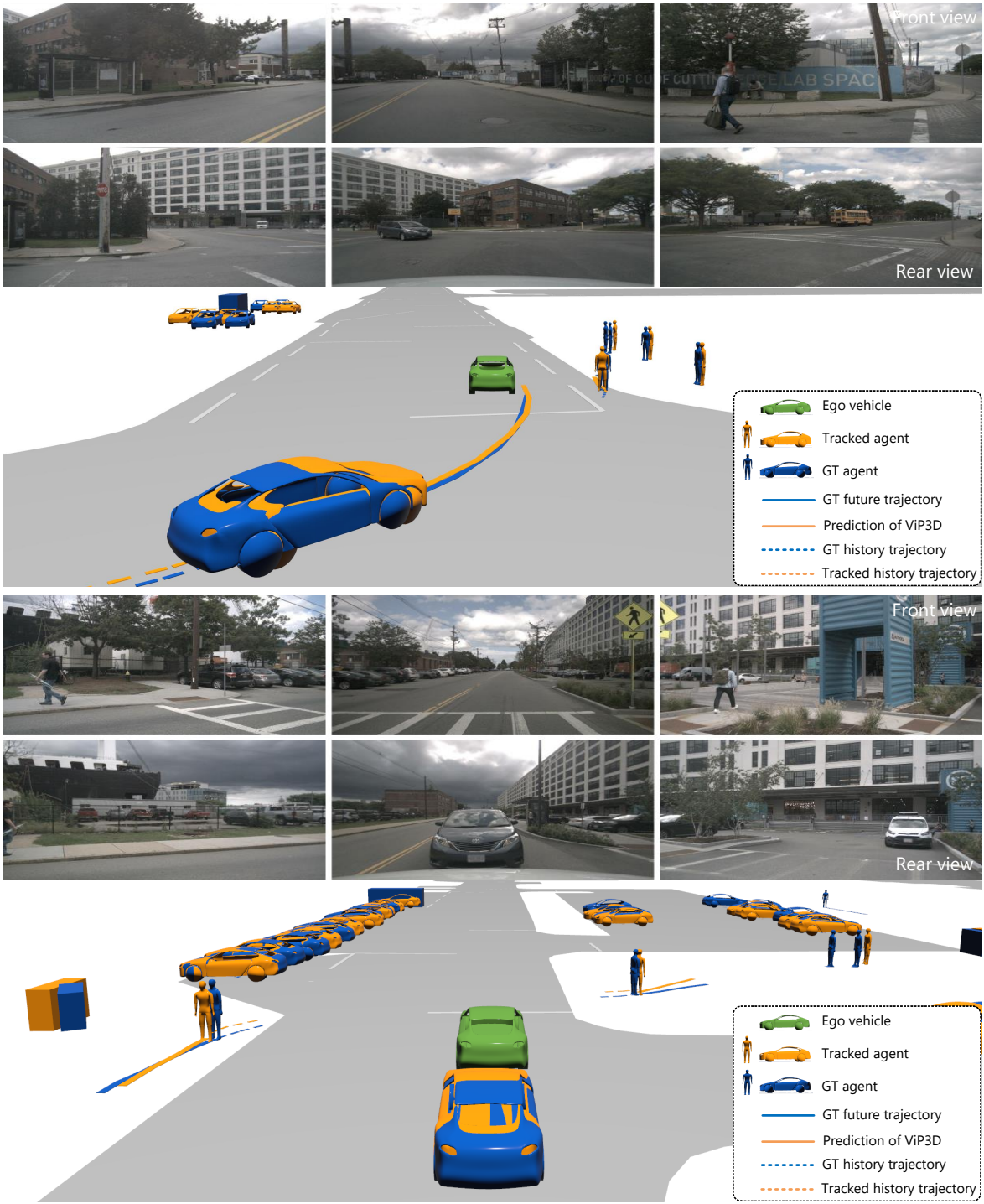}
    \caption{Qualitative results of \model\ on the nuScenes validation set. Input camera images are shown on the top. The green vehicle is the ego agent. The blue and orange agents indicate ground-truth and tracked agents, respectively. The blue and orange curves indicate ground-truth trajectories and predicted trajectories of ViP3D, respectively. For each agent, only the predicted trajectory with the highest probability is drawn.}
    \label{fig:success}
\end{figure*}

\begin{figure*}[!b]
    \centering
    \includegraphics[width=0.9\linewidth]{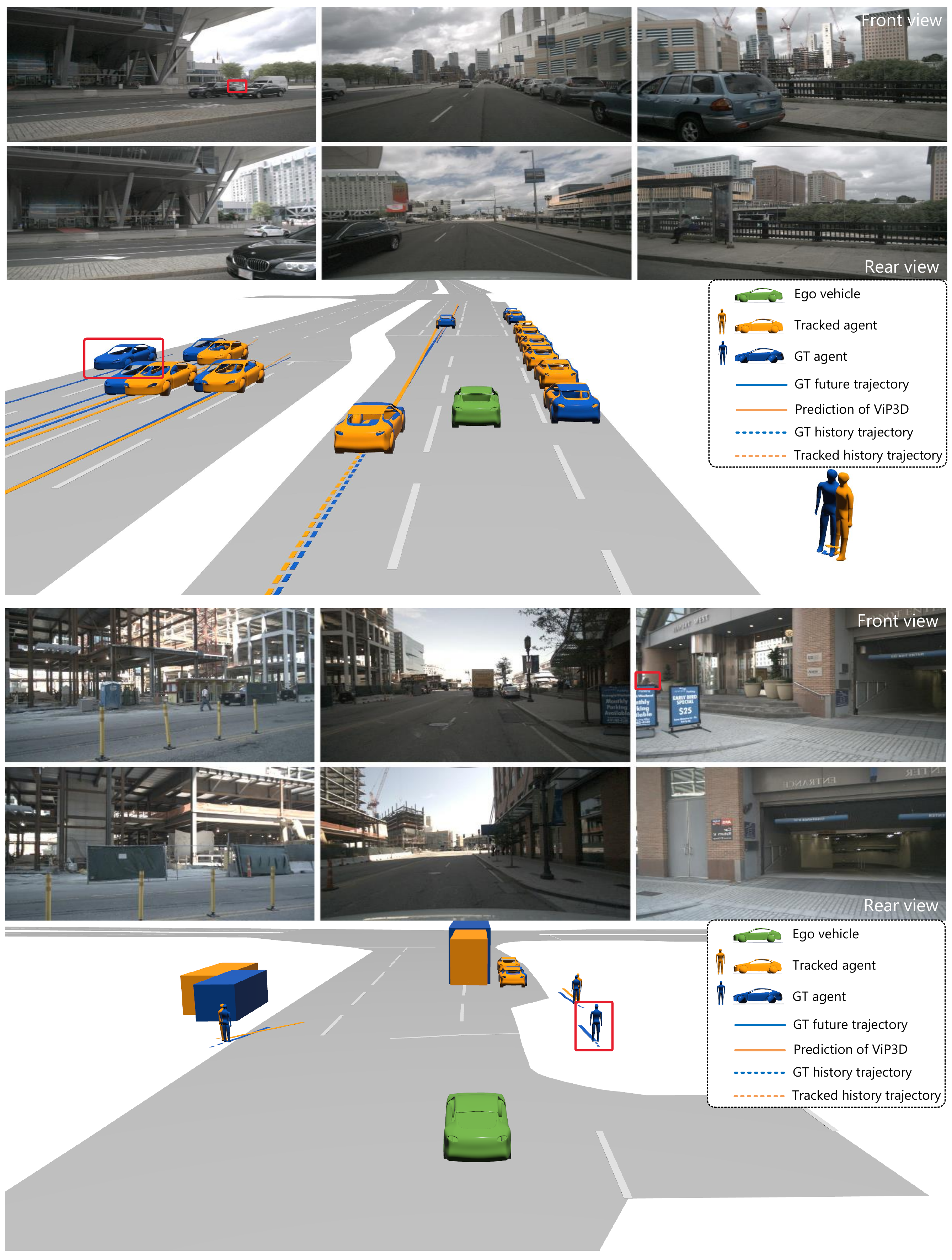}
    \caption{Failure cases of \model\ on the nuScenes validation set. Input camera images are shown on the top. The green vehicle is the ego agent. The blue and orange agents indicate ground-truth and tracked agents, respectively. The blue and orange curves indicate ground-truth trajectories and predicted trajectories of ViP3D, respectively. For each agent, only the predicted trajectory with the highest probability is drawn. The agent surrounded by a red box indicates that it is not detected by ViP3D.}
    \label{fig:failure_1}
\end{figure*}

The visualizations of predicted results of both \model\ and the traditional pipeline are included in the paper. In this section, we provide more visualizations for \model, including some failure cases. As the cases shown in Figure~\ref{fig:success}, \model\ can predict accurate future trajectories. As the failure cases shown in Figure~\ref{fig:failure_1}, because \model\ is a vision-based pipeline, it is difficult for \model\ to detect agents far away from the ego vehicle or agents partially obscured. In the upper part of Figure~\ref{fig:failure_1}, a vehicle (surrounded by a red box) that is far away from the ego vehicle and is partially obscured by other vehicles, so it is difficult to be detected. In the lower part of Figure~\ref{fig:failure_1}, a pedestrian (surrounded by a red box) is mostly obscured by a billboard, so \model\ can not detect this pedestrian.

\clearpage

\end{document}